\documentclass[twoside,11pt]{article}

\usepackage{blindtext}

\usepackage{jmlr2e}
\usepackage{tikz}

\usepackage{hyperref}
\usepackage{listings}
\usepackage{xcolor}
\usepackage{natbib}
\usepackage{sidecap}
\usepackage{caption}

\lstdefinestyle{python}{
   language=Python,
   basicstyle=\ttfamily\small,
   keywordstyle=\color{blue}, stringstyle=\color{green!60!black}, commentstyle=\color{gray},
   showstringspaces=false,
   frame=single,
   breaklines=true
}

\usepackage{lastpage}
\ShortHeadings{CausationEntropy: Pythonic Optimal Causation Entropy}{Slote, Fish and Bollt}
\firstpageno{1}

\begin{document}

\title{CausationEntropy: Pythonic Optimal Causation Entropy for Causal Network Discovery}

\author{\name Kevin Slote \email kslote@clarkson.edu
\AND
\name Jeremie Fish \email jafish@clarkson.edu
\AND
Erik Bollt \email ebollt@clarkson.edu \\
\addr Center of Complex Systems Science\\
Clarkson University\\
Potsdam, NY 13699, USA
}

\editor{Editor}

\maketitle

\begin{abstract}
Optimal Causation Entropy (oCSE) is a robust causal network modeling technique that reveals causal networks from dynamical systems and coupled oscillators, distinguishing direct from indirect paths. \textit{CausationEntropy} is a Python package that implements oCSE and several of its significant optimizations and methodological extensions. In this paper, we introduce the version 1.1 release of CausationEntropy, which includes new synthetic data generators, plotting tools, and several advanced information-theoretical causal network discovery algorithms with criteria for estimating Gaussian, k-nearest neighbors (kNN), geometric k-nearest neighbors (geometric-kNN), kernel density (KDE) and Poisson entropic estimators. The package is easy to install from the PyPi software repository, is thoroughly documented, supplemented with extensive code examples, and is modularly structured to support future additions. The entire codebase is released under the MIT license and is available on \href{https://github.com/Center-For-Complex-Systems-Science/causationentropy}{GitHub} and through PyPi Repository. We expect this package to serve as a benchmark tool for causal discovery in complex dynamical systems.
\end{abstract}%


\section{Introduction}

Cause-and-effect relationships are a central interest of scientists and researchers in science and engineering, where the discovery of causal networks from dynamical systems and coupled oscillators is critical for explanation, prediction, and control. The optimal causation entropy principle states that the causal parents of a network time series form the minimal set of nodes that maximize the causation entropy \citep{sun2015causal, Sun2014CausationEntropy}. Researchers have developed several algorithms for the discovery of causal networks, including widely used approaches such as PCMCI and its variants such as PCMCI+, LPCMCI and RPCMCI \citep{Runge2019PCMCI, Runge2020PCMCIPlus, lpcmci, Saggioro2020RPCMCI, Runge2018Overview}--as well as transfer entropy \citep{schreiber2000measuring} and traditional Granger causality \citep{Granger1969, Dhamala2018}. The \textit{tigramite} library, an implementation of the PCMCI algorithm, has been applied in a wide range of studies in many disciplines (e.g. \citep{slote2025, Tarraga2024CausalDroughtDisplacement, Ganesh2023}), facilitated by its open-source availability. In contrast, comparable implementations of optimal causation entropy have not been readily accessible, motivating our development of an open-source implementation. In this work, we introduce an improved API that reduces the barrier to entry for causal inference using oCSE. In particular, we demonstrate that CausationEntropy offers a principled and versatile alternative.

Optimal Causation Entropy applications exist in a wide range of domains, including neuroscience \citep{McIntyre2024Contrasting, fish2021}, nonlinear dynamical system model discovery \citep{AlMomani2020}, and parameter estimation \citep{kim2017causation}. These studies highlight oCSE's detection of direct and indirect causal drivers in dynamical systems while avoiding spurious causal relationships. The information-theoretical basis of oCSE ensures that inferred causal structures are statistically principled and interpretable within the context of dynamical system analysis. Consequently, the growing literature employs oCSE demonstrating both its generality and its effectiveness in finding hidden network structures in complex multivariate data.

Previous implementations of oCSE are largely research-specific prototypes written in low-level code, lacking modularity, and providing little open-source ecosystem availability. Our Pythonic implementation addresses these limitations directly: integrates oCSE and its methodological extensions into a cohesive, well-documented package with extensive examples, testing infrastructure that includes 354 unit tests were generated and verified with the assistance of large language model code review tools \citep{Claude2025, Gemini2025} and vetting one by one yielding 100\% test coverage accounting for every possible numerical calculation and code path, and visualization tools. Code modularity enables easy code open-source contributions with the incorporation of new entropy estimators, while a consistent API lowers the barrier to adoption for practitioners without a background in information theory. The open-source library is released under the MIT license and is maintained openly on \href{https://github.com/Center-For-Complex-Systems-Science/causationentropy}{GitHub}.

\begin{figure}[!htp]
\centering
\begin{tikzpicture}
% Draw circles
\draw[thick] (0,0) circle(1.5cm) node[left=2.2cm, above=1.5cm] {\small$H(I_{t+1})$};
\draw[thick] (1.5,0) circle(1.5cm) node[right=1.2cm, above=1.5cm] {\small$H(J_t)$};
\draw[thick] (0.75,-1.4) circle(1.5cm) node[below right=1.6cm] {\small$H(K_t)$};
% Labels for sets
\node at (-0.6,0.2) {$I_{t+1}$};
\node at (2.1,0.2) {$J_t$};
\node at (0.75,-2.4) {$K_t$};
% Overlap labels
\node at (-2.50,-1.0) {\small$H(I_{t+1},J_t)$};
% Shaded causation entropy region
\begin{scope}
    \clip (0,0) circle(1.5cm);
    \clip (1.5,0) circle(1.5cm);
    \fill[blue!30, even odd rule] (-5,-5) rectangle (5,5) (0.75,-1.4) circle(1.5cm);
\end{scope}
% Legend for causation entropy
\fill[blue!30] (3.5,0.5) rectangle +(0.3,0.3);
\node[anchor=west] at (3.9,0.65) {\small causation entropy};
\node[anchor=west] at (3.9,0.25) {\small $C_{J \to I \mid K}$};
% Text about special case
\node[anchor=west] at (3.5,-1.1) {\small ($K = I \Rightarrow C_{J \to I \mid K} = T_{J \to I}$)};
\end{tikzpicture}
\caption{A Venn diagram illustrating causation entropy and its relationship to transfer entropy. }
\label{fig:venn}
\end{figure}
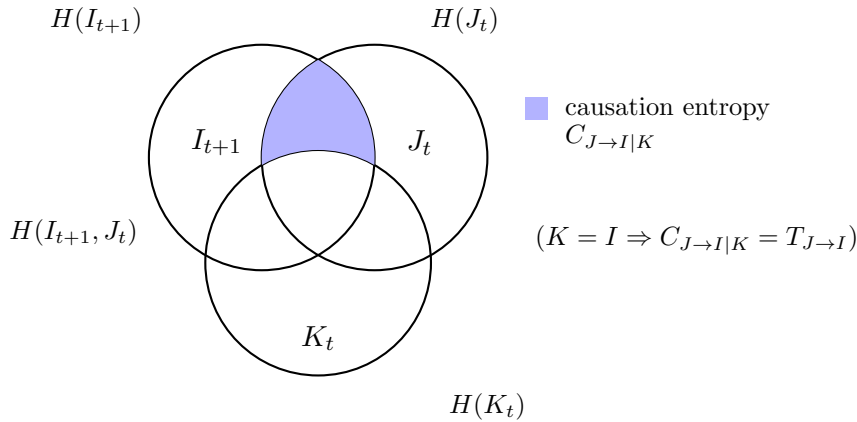

Figure~\ref{fig:venn} shows the Venn diagram of the information-theoretical principles of oCSE. The oCSE method consists of two steps: first, a forward pass to determine the minimal set of cause parents that maximizes the conditional mutual information, and then a backward pass, which prunes the network to identify the actual causal structure. The CausationEntropy library itself allows both significance thresholds for forward and backward passes, with a default of $0.05$. The shuffle test determines the significance thresholds, which the user can specify and which default to $200$. The algorithm computes the causation entropy by estimating the conditional mutual information for the lagged variables of the causal parents in both forward and backward directions. Transfer Entropy \citep{schreiber2000measuring} represents a special case of causation entropy when no conditioning set of causal parents corrects the information measure. Conditional mutual information is an additional parameter for the module. The available information-theoretic quantities are Gaussian, kNN, geometric-kNN \citep{Lord2018gknn}, and Poisson data \citep{Fish2022PoissonMMI}. The modular design of the library enables easy extension with new information-theoretic quantities; for example, a new information criterion requires updating the entropy, mutual information, and conditional mutual information calculations in the library's information module. An easy how-to guide can be found in \href{https://github.com/Center-For-Complex-Systems-Science/causationentropy/tree/main/notebooks}{GitHub}.

\section{CausationEntropy Structure and Usage}

The CausationEntropy package is organized into modular subpackages that separate core information-theoretic routines, causal network discovery algorithms, and utilities for data handling and visualization. At its foundation, the \texttt{core.information} module implements efficient estimators of entropy, mutual information, and conditional mutual information, which serve as building blocks for oCSE-based inference. The \texttt{core} module provides the main discovery routines, including both forward and backward oCSE selection procedures, along with recent extensions that handle multiscale dynamics, high-dimensional embedding, and nonlinear dependencies. To support real-world applications, the package also includes \texttt{datasets} modules for generating synthetic time series of coupled oscillators, as well as plotting tools for visualizing causal graphs, adjacency matrices, and temporal dynamics.

From the user's perspective, the workflow is intentionally streamlined. After importing the package, a typical analysis begins by passing a multivariate time series to the network discovery function, \verb|discover_network|. Users can configure the entropy estimator (e.g., Gaussian, kNN), adjust statistical thresholds, and select from multiple optimization criteria. The output is a directed multigraph Networkx \citep{hagberg2008exploring} object with directed edges for the corresponding lags where the oCSE detected a significant causation entropy, and each edge comes with an additional attribute of the p-value of the shuffle test and the conditional mutual information for that edge,  which can be further analyzed, visualized or exported. Extensive tutorials and example notebooks are provided in the public documentation to demonstrate use cases ranging from small synthetic networks to large, noisy experimental datasets. This combination of modular architecture, consistent API design, and accessible examples ensures that both domain scientists and methodological researchers can easily adopt, extend, and benchmark the package in their own work.

\newpage
\begin{lstlisting}[style=python, caption={Network discovery example}, label={lst:discover_network}]
from causationentropy import discover_network
from causationentropy.datasets import synthetic

data, true_network = synthetic.linear_stochastic_gaussian_process(
    n=5, 
    T=1000, 
    rho=0.7,
    p=0.2
)

network = discover_network(data)
\end{lstlisting}

As shown in Code Sample~\ref{lst:discover_network}, the causal network is discovered automatically. The \verb|plot_causal_network| routine can then be used to visualize the components of the discovered network. This function accepts a variety of plotting parameters and can be used to generate figures such as Figure 1. For more examples highlighting specific oCSE variants and use cases, see our complete set of Jupyter Notebook tutorials \href{https://github.com/Center-For-Complex-Systems-Science/causationentropy/tree/main/notebooks}{GitHub}.

\begin{SCfigure}[][ht]
    \centering
    \includegraphics[width=0.5\linewidth]{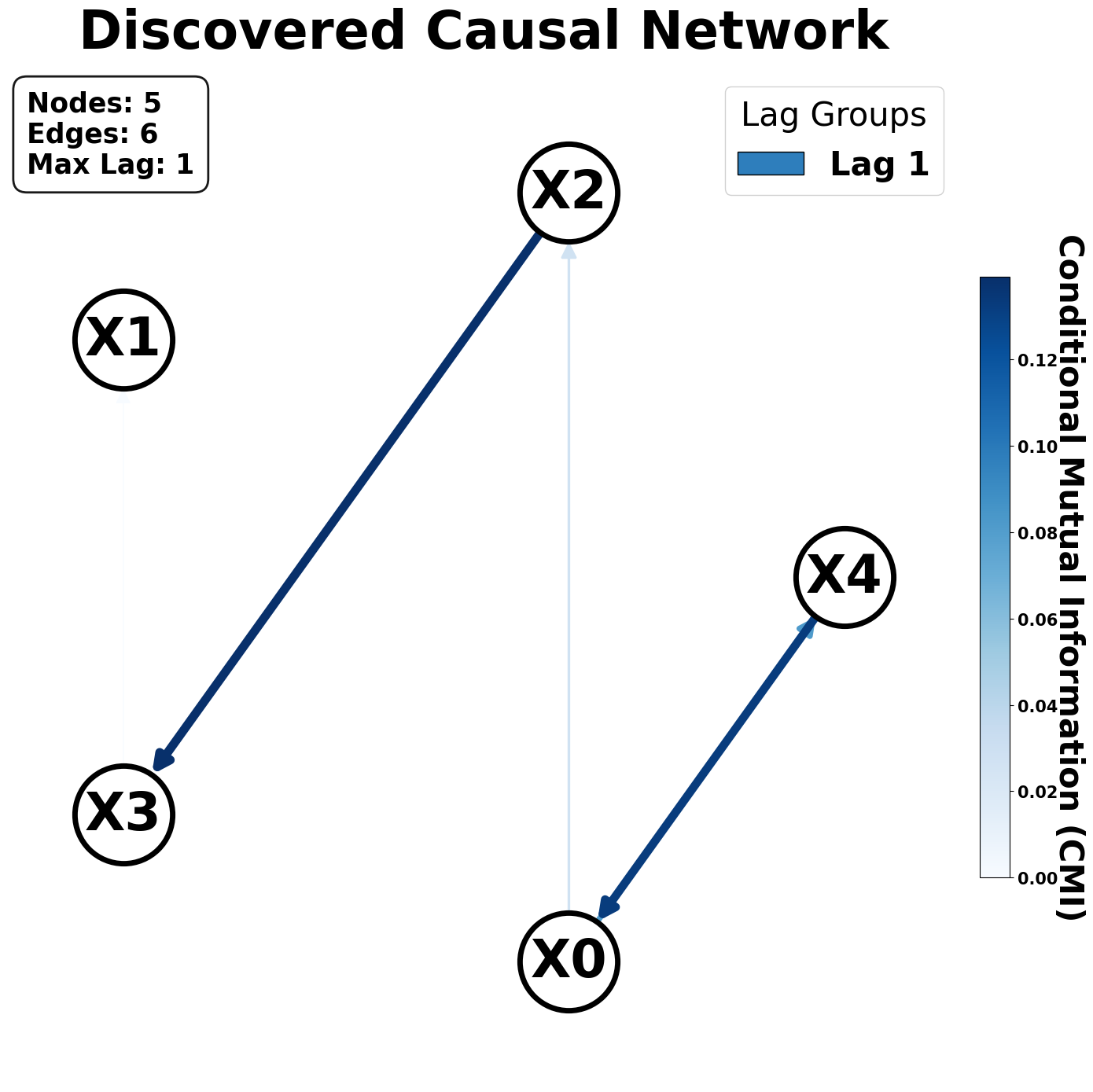}
    \captionsetup{labelformat=empty}
    \caption{Figure~\ref{fig:plot_network}. Causal network discovered with the metwork discovery method for coupled Gaussian oscillators on an Erd\H{o}s-R\'enyi random graph with edge probability of $0.2$ and coupling strength of $\rho=0.7$ with $5$ nodes plotted with the method to plot a general causal network. The network discovery method, using standard forward and backward OSE methods and the Gaussian information criterion, identified the correct causal network. The directed edges indicate statistically significant causal influences detected by the shuffle test at $\alpha = 0.05$. Edge color corresponds to the magnitude of the conditional mutual information, and arrow direction denotes the inferred direction of information flow.}
    \label{fig:plot_network}
\end{SCfigure}

Figure~\ref{fig:plot_network} shows the visualized causal network returned by the \verb|plot_causal_network| function for a random graph Erd\H{o}s-R\'enyi used to generate the coupling of the oscillators. Other available methods include \verb|network_to_dataframe|, which allows users to examine the algorithm output as a Pandas data frame with columns for Source, Sink, CMI, and P-value.

\section{Conclusion}
The CausationEntropy package is an open-source project that enables users with diverse mathematical backgrounds to apply optimal Causation Entropy within a user-friendly Pythonic environment. Our latest updates, featured in CausationEntropy version 1.1, additionally make it easier than ever for users to use and visualize results from state-of-the-art causal network discovery methods that are capable of extracting coherent causal structures from real-world datasets. Through this work and future endeavors like this, CausationEntropy can continue to serve as a practical data analysis tool and as an ever-expanding centralized codebase for causation entropy methods.

\acks{This work was supported by the Army Research Office under Grant No.~W911NF2310393.
We also thank Kelley Smith for being an early adopter and contributor to our software.}

\vskip 0.2in

\bibliography{references}

\end{document}